\documentclass[10pt,twocolumn,letterpaper]{article}
\usepackage[affil-it]{authblk}
\usepackage[pagenumbers]{cvpr} 

\definecolor{cvprblue}{rgb}{0.21,0.49,0.74}
\usepackage[pagebackref,breaklinks,colorlinks,allcolors=cvprblue]{hyperref}

\newcommand\blfootnote[1]{%
  \begingroup
  \renewcommand\thefootnote{}\footnote{#1}%
  \addtocounter{footnote}{-1}%
  \endgroup
}

\def\paperID{13} 
\def\confName{CVPR}
\def\confYear{2025}

\title{A Training-Free, Task-Agnostic Framework for Enhancing MLLM Performance on High-Resolution Images}

\author{Jaeseong Lee*, 
Yeeun Choi*, 
Heechan Choi*,
Hanjung Kim,
Seonjoo Kim
}
\affil{Yonsei University}

\usepackage{graphicx}
\usepackage{booktabs}
\usepackage{multirow, arydshln}
\usepackage{adjustbox}
\usepackage{array}
\begin{document}
\maketitle
\blfootnote{* indicates equal contribution.}

\begin{abstract}

Multimodal Large Language Models (MLLMs) have demonstrated remarkable capabilities in vision-language understanding, reasoning, and generation. However, they struggle with tasks requiring fine-grained localization and reasoning in high-resolution images. This constraint stems from the fact that MLLMs are fine-tuned with fixed image resolution to align with the pre-trained image encoder used in MLLM. Consequently, feeding high-resolution images directly into MLLMs leads to poor generalization due to a train-test resolution discrepancy, while downsampling these images—although ensuring consistency—compromises fine-grained visual details and ultimately degrades performance. To address this challenge, we propose \textbf{E}xtract \textbf{C}andidate then \textbf{P}redict (\textbf{ECP}), a novel training-free, task-agnostic two-stage framework designed to enhance MLLM performance on high-resolution images. The key intuition behind \textbf{ECP} is that while MLLMs struggle with high-resolution images, their predictions on downsampled images still contain implicit localization cues. By first identifying candidate region using the coarse prediction and then predicting the final output based on candidate region, \textbf{ECP} effectively preserves fine-grained details while mitigating the challenges posed by high-resolution data. We validate our framework on 4K GUI grounding and 4K, 8K MLLM perception, achieving +21.3\%, +5.8\%, +5.2\% absolute improvement compared to baseline respectively, demonstrating its effectiveness. Code is available at \url{https://github.com/yenncye/ECP}.

\end{abstract}
    
\section{Introduction}
\label{sec:intro}

\begin{figure}
    \centering
    \includegraphics[width=\linewidth]{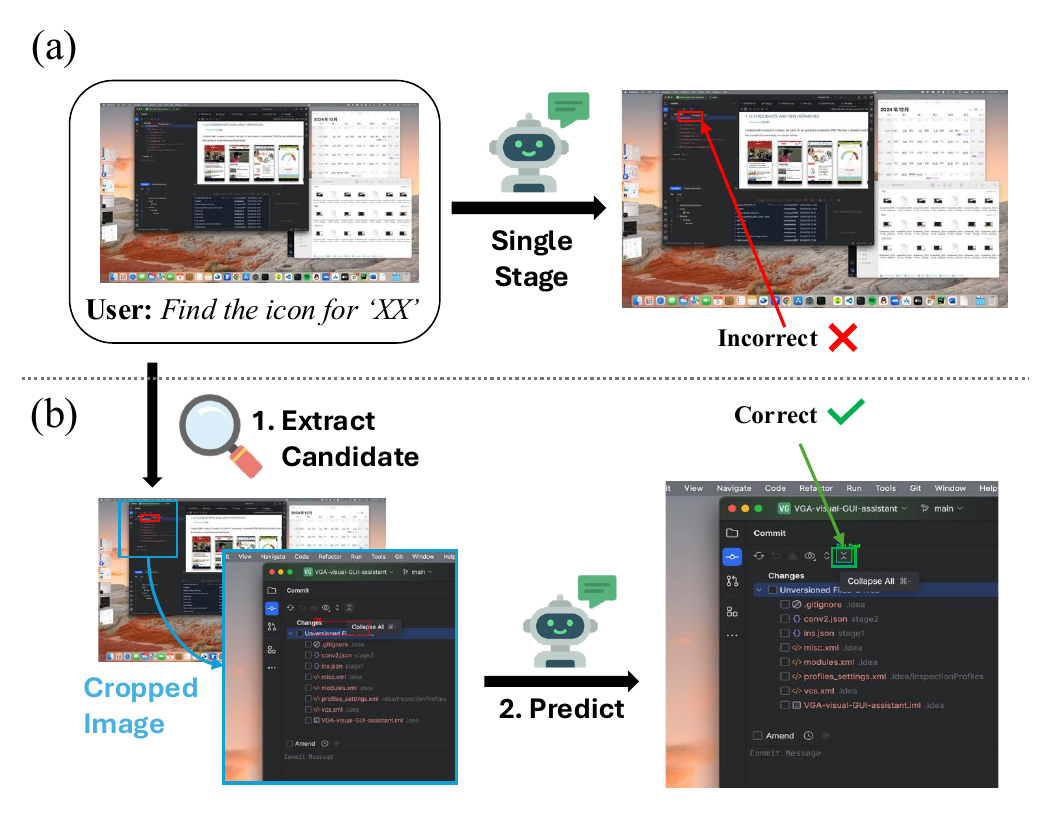}
    \caption{\textbf{(a)} Overview of the conventional single-stage framework, where a high-resolution image is first downsampled, and the MLLM generates an output based on downsampled image and text instruction. \textbf{(b)} Overview of our proposed \textbf{E}xtract \textbf{C}andidate then \textbf{P}redict (\textbf{ECP}) framework. In the first stage, the MLLM processes a downsampled high-resolution image and generates a candidate region in the form of either a point or a bounding box. In the second stage, the high-resolution image is cropped based on the output from the MLLM. The resulting cropped patch is then used for the final prediction, which enables the extraction of fine-grained details from the image.}
    \label{fig:fig1}
    \vspace{-5mm}
\end{figure}

Multimodal Large Language Models (MLLMs)~\cite{llava, qwenvl, internvl} have demonstrated strong capabilities in understanding, reasoning, and generation, achieving remarkable success across vision, language, and vision-language tasks. Leveraging these advancements, numerous MLLM-powered products~\cite{gpt4o, gemini15} have emerged, transforming industries and enhancing human efficiency.

Despite their impressive capabilities, MLLMs struggle with tasks requiring precise localization and reasoning in high-resolution images~\cite{omgllava, screenspot-pro, hrbench, mrovseg}. This limitation arises because MLLMs are fine-tuned with fixed image resolution~\cite{llava, qwenvl, internvl} to maintain consistency with a pre-trained image encoder used in MLLM. As a result, directly feeding high-resolution images into MLLMs leads to poor generalization~\cite{fixres}. Downsampling high-resolution images ensures consistency but causes a loss of fine-grained visual information, impairing the model’s ability to capture the details of high-resolution images.

To address this challenge, prior research has explored two main directions. The first direction is adding specialized modules trained on high-resolution datasets~\cite{llava-uhd, llava-hr, infimm-hd, inf-llava}. However, this approach is inefficient, requiring manual module design, dataset construction, and extensive computation resources. The second direction is employing training-free, task-specific two-stage framework~\cite{hrbench, dc}, where high-resolution images are divided into patches, and per-patch predictions are combined using score averaging~\cite{dc} or tree-like structure~\cite{hrbench}. While this approach removes the need for additional training, it still relies on heuristic designs tailored to specific task, limiting its generalizability. This raises a fundamental question: can we make a training-free, task agnostic framework that eliminates the need for manual task-specific design?

Based on empirical observations, we assume that although MLLMs struggle with high-resolution images, their predictions on downsampled images exhibit an implicit understanding of where to attend. This suggests that despite performance degradation, MLLMs retain a coarse localization ability even when fine-grained details are lost due to downsampling.

By leveraging this insight, we propose \textbf{E}xtract \textbf{C}andidate then \textbf{P}redict (\textbf{ECP}), a novel training-free, task-agnostic two-stage framework designed to enhance MLLM performance on high-resolution images. In the first stage, the \textbf{E}xtract \textbf{C}andidate (\textbf{EC}) generates instruction-relevant candidate region by feeding a downsampled version of the high-resolution image into MLLMs. In the second stage, the \textbf{P}redict (\textbf{P}) makes the final prediction based on the extracted candidate region. By decomposing high-resolution image understanding into a coarse-to-fine process, \textbf{ECP} effectively mitigates the challenges posed by high-resolution data while preserving fine-grained details for accurate predictions.

We validate our \textbf{ECP} framework on two applications: 4K GUI grounding~\cite{screenspot-pro} and 4K, 8K MLLM perception~\cite{hrbench}. With its efficient design, our framework achieves +21.3\%, +5.8\%, +5.2\% absolute improvement compared to baseline on 4K GUI grounding, 4K, 8K MLLM perception respectively, demonstrating its effectiveness in improving MLLM capabilities for high-resolution image understanding.

In summary, our key contributions are summarized as follows:

1. We propose \textbf{E}xtract \textbf{C}andidate then \textbf{P}redict (\textbf{ECP}), a novel training-free, task-agnostic two-stage framework that enhances MLLMs performance on high-resolution images.

2. We validate our \textbf{ECP} framework on 4K GUI grounding and 4K, 8K MLLM perception, achieving strong results and demonstrating its effectiveness.
\section{Related Works}
\label{sec:Related}

\noindent \textbf{Multimodal Large Language Models.}\quad
Large Language Models (LLMs)~\cite{gpt3, qwen, llama, palm} have demonstrated remarkable capabilities in language understanding, reasoning, and generation, leading to a paradigm shift in natural language processing (NLP). To extend these powerful capabilities to vision-language tasks, Multimodal Large Language Models (MLLMs)~\cite{gpt4o, gemini15, qwenvl, internvl, llava} have emerged, integrating vision and language processing within a unified framework.

MLLMs typically consist of three primary components: an image encoder, a connector, and an LLM. The image encoder extracts visual features from an image input. The connector then projects these visual features into a language space that the LLM can interpret. Finally, the LLM performs tasks by leveraging the projected visual features along with text instructions.

\noindent \textbf{Processing High-Resolution Images.}\quad
Most MLLMs downsample input images to a fixed resolution to match the input requirements of pre-trained image encoders~\cite{clip, dino}, which limits their ability to handle fine-grained reasoning in high-resolution images. To address the challenge of fine-grained reasoning in high-resolution images, prior works~\cite{llava-uhd, llava-hr, infimm-hd, inf-llava} have introduced specialized modules for high-resolution image processing.
LLaVA-UHD~\cite{llava-uhd} introduces a Modularized Visual Encoding strategy that adaptively slices high-resolution images into variable-sized regions, followed by a Compression Layer that reduces visual tokens before integrating them into the language model with spatial encoding.
Another line of work, LLaVA-HR~\cite{llava-hr} introduces Dual Visual Pathways to process both low- and high-resolution inputs, and employs a Mixture-of-Resolution Adapter to fuse fine-grained features into the low-resolution stream efficiently.
Similarly, INF-LLaVA~\cite{inf-llava} separately encodes local and global crops and combines them using dual-perspective cropping and enhancement module.
However, these approaches are inefficient, as they require additional model training with high-resolution image datasets. This highlights the need for a training-free method that leverages pre-trained MLLMs without additional data or fine-tuning.

Other works~\cite{hrbench, dc} attempt to address high-resolution input by dividing images into patches and aggregating their predictions. Yet, these methods still depend on manually crafted heuristics tailored to specific tasks, which limits their generalizability. Similar to our approach, \cite{screenspot-pro} explores iterative prompting—where the model first produces a coarse prediction of the answer location and then progressively zooms in for refinement—but this method remains restricted to GUI grounding tasks. In contrast, our work introduces a task-agnostic, training-free two-stage framework applicable to diverse high-resolution tasks, and demonstrates its effectiveness across multiple benchmarks.
\begin{figure*}
    \centering
    \includegraphics[width=\linewidth]{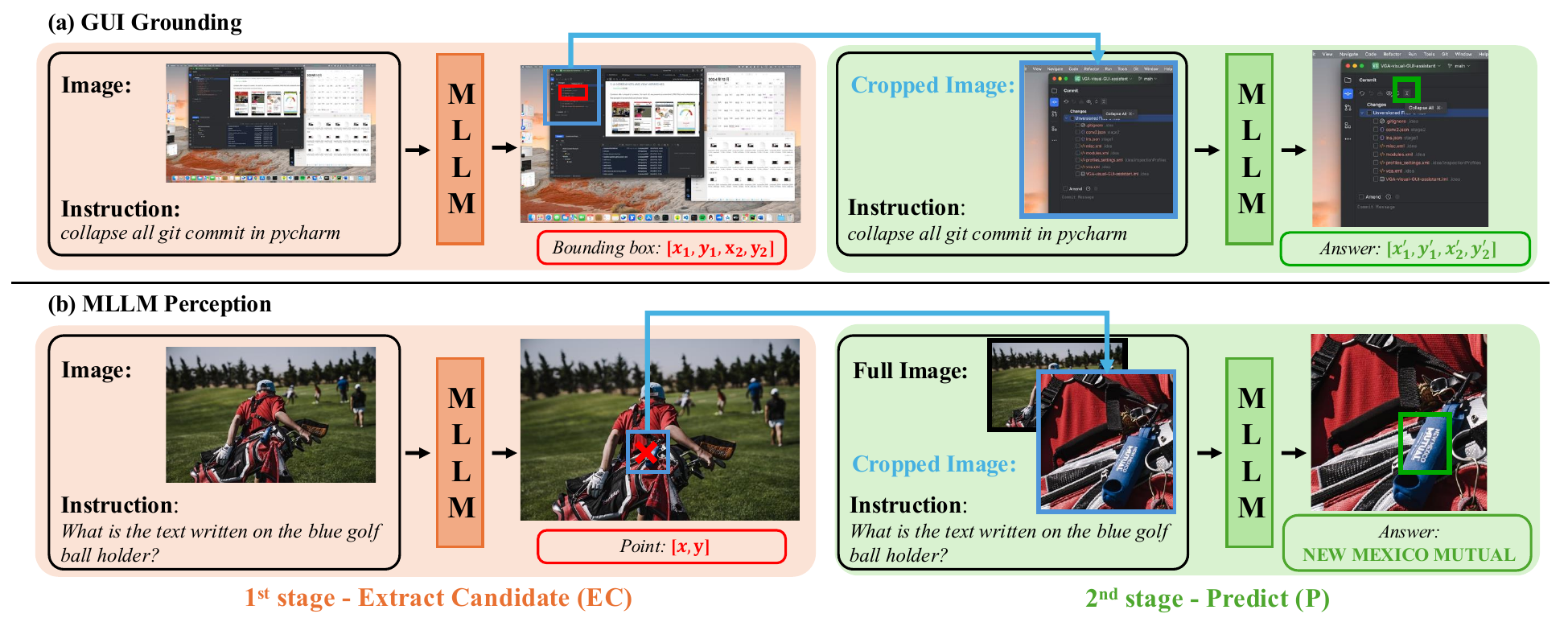}
    \caption{\textbf{(a)} Overview of the \textbf{ECP} framework in GUI Grounding. In the first stage, \textbf{E}xtract \textbf{C}andidate (\textbf{EC}), MLLM processes a downsampled high-resolution image and text instruction to generate bounding box for candidate extraction. In the second stage, \textbf{P}redict (\textbf{P}), MLLM generates final prediction based on candidate region and text instruction.
    \textbf{(b)} Overview of \textbf{ECP} framework in MLLM Perception. In \textbf{EC}, MLLM generates instruction-relevant point given a downsampled high-resolution image and text instruction. In \textbf{P}, MLLM makes final prediction based on downsampled high-resolution image, cropped candidate region, and text instruction.}
    \label{fig:fig2}
    \vspace{-5mm}
\end{figure*}
\section{Method}

We propose a two-stage framework designed for high-resolution image understanding with MLLMs, where the first stage identifies instruction-relevant region, and the second stage makes final prediction based on extracted region.

\subsection{Stage 1: Extract Candidate (EC)}

While MLLMs struggle with precise localization and reasoning in high-resolution images, we assume that they can still identify which area of the image should be focused on. \textbf{E}xtract \textbf{C}andidate (\textbf{EC}) does not directly generate an answer to user's instruction for a high-resolution image. Instead, it finds the instruction-relevant region in the image, represented as a bounding box, that should be focused on to generate a response for the instruction.


To achieve this, we first leverage a pre-trained MLLM, denoted as \( \mathcal{F}_{EC} \), that takes a high-resolution RGB image with width \( W \) and height \( H \) along with a text instruction as input and then outputs either a point or a bounding box:
\begin{equation}
O_{point} = (x, y), \quad O_{bbox} = (x_1, y_1, x_2, y_2)
\end{equation}
where \( O_{point} \) denotes a single point of interest, and \( O_{bbox} \) denotes a bounding box with top-left coordinate \( (x_1, y_1) \) and bottom-right coordinate \( (x_2, y_2) \). We assume that \( \mathcal{F}_{EC} \) applies downsampling when processing input images.


Depending on the type of output from $\mathcal{F}_{EC}$, we define a representative coordinate as follows: if the output is a point $O_{{point}}$, we use it directly; if it is a bounding box $O_{{bbox}}$, we compute its center. We denote this representative coordinate as $O_{{rep}} = (x_{{rep}}, y_{{rep}})$.

Finally, we compute the instruction-relevant region as a bounding box \( B_{candidate} = (x_{\text{left}}, y_{\text{top}}, x_{\text{right}}, y_{\text{bottom}}) \). This bounding box is determined by using \( O_{rep} \) to define its center, and the hyperparameters \( w \), and \( h \) to set its width and height, respectively, where \( w < W, h < H \). In our experiments, we set \( w=1024, h=1024 \). To ensure that the \( B_{candidate} \) remains within the image boundaries, we adjust as follow:
\begin{equation}
x_{\text{left}} = \max(0, \min(x_{\text{rep}} - w/2, W-w))
\end{equation}
\begin{equation}
y_{\text{top}} = \max(0, \min(y_{\text{rep}} - h/2, H-h))
\end{equation}
\begin{equation}
x_{\text{right}} = x_{\text{left}} + w, \quad y_{\text{bottom}} = y_{\text{top}} + h
\end{equation}

\subsection{Stage 2: Predict (P)}

\textbf{P}redict (\textbf{P}) generates the final prediction by focusing on the instruction-relevant region $B_{candidate}$ identified in the previous stage. By feeding the cropped patch of the high-resolution image around $B_{candidate}$, MLLMs are able to capture fine-grained details (see Fig.~\ref{fig:fig2} (a)). When global context is also essential, such as MLLM perception task, we supplement the cropped patch with a downsampled high-resolution image along with the user's instruction (see Fig.~\ref{fig:fig2} (b)).






\section{Experiments}
\label{sec:Experiments}
We compare the conventional single-stage framework with our proposed two-stage framework, \textbf{ECP}, on two challenging tasks: 4K GUI grounding and 4K, 8K MLLM perception. As shown in Fig.~\ref{fig:fig1} (a), the the single-stage framework predicts output based on downsampled high-resolution image with text instruction. In contrast, our \textbf{ECP} framework follows a two-stage process illustrated in Fig.~\ref{fig:fig2}.

To highlight the MLLM's coarse localization ability to generate instruction-relevant region in \textbf{EC}, we perform an ablation study using a random sampling in \textbf{EC}, instead of using \( \mathcal{F}_{EC} \), where \( O_{rep} \) is selected uniformly across the entire image.

\subsection{GUI Grounding}

\subsubsection{Experiment Setting}

\noindent \textbf{Benchmark}\quad
We evaluate our framework on ScreenSpot-Pro \cite{screenspot-pro}, a GUI grounding benchmark with images at resolution 4K, designed to overcome the limitations of previous benchmarks~\cite{seeclick, os-atlas}. ScreenSpot-Pro includes six application categories, primarily focusing on development (Dev), creative software (Cre), CAD/engineering (CAD), and scientific analysis (Sci). It also supports office productivity tools (Office) and common operating system tasks (OS), covering document processing, data management, and system utilities.

\noindent \textbf{Evaluation Metric}\quad
Following the previous works~\cite{seeclick, os-atlas, screenspot-pro}, we use accuracy, which determines correctness if the predicted point or the center of the predicted bounding box falls within the ground-truth bounding box. 

\noindent \textbf{Models}\quad
In the first stage, \textbf{EC}, we use either Qwen2-VL-7B \cite{qwenvl} or OS-Atlas-7B \cite{os-atlas} as \( \mathcal{F}_{EC} \). For ablation, we add random sampling. In the second stage, \textbf{P}, we use either Qwen2-VL-7B or OS-Atlas-7B.

\newcommand\mr[2]{\multicolumn{1}{c}{\multirow{#1}{*}{\makecell{#2}}}}

\begin{table}[t!]
\centering
\resizebox{\columnwidth}{!}{%
\renewcommand{\arraystretch}{1.2}
\begin{tabular}{ccccccccc}
\toprule
\multicolumn{2}{c}{\textbf{Models}} & \textbf{Dev} & \textbf{Cre} & \textbf{CAD} & \textbf{Sci} & \textbf{Office} & \textbf{OS} & \textbf{Overall}
\\ 
\midrule
\multicolumn{9}{l}{\cellcolor[gray]{0.9} \textbf{\textit{Single-stage Framework}}}  \\ 
\hline
    \multicolumn{2}{c}{GPT-4o* \cite{gpt4o}} & 0.7 & 0.6 & 1.5 & 1.2 & 0.9 & 0.0 & 0.8 \\ 
    \hline
    \multicolumn{2}{c}{Qwen2-VL-7B \cite{qwenvl}}  & 1.7 & 1.2 & 0.4 & 3.1 & 2.6 & 0.0 & 1.5 \\ 
\hline
    \multicolumn{2}{c}{OS-Atlas-7B \cite{os-atlas}} & \textbf{18.7} & \textbf{19.4} & \textbf{9.2} & \textbf{23.2} & \textbf{27.8} & \textbf{16.8} & \textbf{19.1} \\ 

\midrule

\multicolumn{9}{l}{\cellcolor[gray]{0.9} \textbf{\textit{ECP Framework (Ours)}}}  \\ 
\hline
\multirow{2}{*}{Random Sampling} 
& Qwen2-VL-7B 
    & 2.7 & 5.9 & 5.0 & 10.6 & 5.7 & 2.6 & 5.4 \\
& OS-Atlas-7B
  & 7.4 & 12.6 & 11.1 & 18.5 & 13.5 & 8.2 & 11.9 \\ 
 \hline
\multirow{2}{*}{Qwen2-VL-7B}
& Qwen2-VL-7B 
    & 8.4 & 9.7 & 4.6 & 16.1 & 20.4 & 11.7 & 11.4 \\
& OS-Atlas-7B
 & 22.4 & 27.9 & 12.3 & 26.8 & 40.9 & 24.5 & 25.6 \\ 
 \hline
\multirow{2}{*}{OS-Atlas-7B}
& Qwen2-VL-7B 
    & 14.4 & 17.3 & 8.8 & 23.6 & 31.3 & 20.4 & 18.8 \\
& OS-Atlas-7B
  & \textbf{38.8} & \textbf{39.0} & \textbf{27.2} & \textbf{40.2} & \textbf{58.3} & \textbf{42.3} & \textbf{40.4} \\ 
\bottomrule
\end{tabular}
}
\caption{Grounding accuracy comparison of different models in the single-stage and \textbf{ECP} frameworks on ScreenSpot-Pro~\cite{screenspot-pro}. The values marked with * are reported from previous works~\cite{screenspot-pro}.}
\label{tab:main_gui}
\vspace{-5mm}
\end{table}

\subsubsection{Experiment Results}
Table~\ref{tab:main_gui} presents the results of the conventional single-stage framework and our \textbf{ECP} framework on ScreenSpot-Pro~\cite{screenspot-pro}. 

\noindent \textbf{Single-stage Framework Results}\quad
In our single-stage framework experiments, we observe that general-purpose MLLMs, GPT-4o~\cite{gpt4o} and Qwen2-VL-7B~\cite{qwenvl}, achieve extremely low performance on high-resolution GUI images, with accuracy of 0.8\% and 1.5\%, respectively. Even OS-Atlas-7B~\cite{os-atlas},  which is a task-specific MLLM designed for GUI grounding, attains 19.1\%. These results confirm that MLLMs struggle with tasks requiring fine-grained localization and reasoning in high-resolution images.

\noindent \textbf{ECP Framework Results}\quad
In contrast, our \textbf{ECP} framework substantially boosts performance across all categories. For instance, when using Qwen2-VL-7B in \textbf{EC}, Qwen2-VL-7B's performance improves 1.5\% to 11.4\% (a 9.9\% increase). Moreover, the \textbf{EC} with OS-Atlas-7B further boosts performance to 18.8\% (a 17.3\% improvement). Similarly, for OS-Atlas-7B in \textbf{P}, integrating Qwen2-VL-7B in \textbf{EC} raises the accuracy from 19.1\% to 25.6\% (a 6.5\% boost), and using OS-Atlas-7B in \textbf{EC} results in an even higher accuracy of 40.4\% (a 21.3\% gain). These consistent improvements highlight that our \textbf{ECP} framework effectively addresses the challenges of high-resolution image tasks. While our ECP framework is designed to be training-free and task-agnostic, selecting a well-suited MLLM for each task can further improve performance.



\noindent \textbf{Ablation}\quad
To validate the importance of instruction-guided region selection in \textbf{EC}, we conducted an ablation study by replacing it with a random sampling. For Qwen2-VL-7B, changing to random sampling degrades performance from 11.4\% to 5.4\% (a drop of 6.0\%) compared to using either Qwen2-VL-7B or OS-Atlas-7B in \textbf{EC}. Likewise, for OS-Atlas-7B in \textbf{P}, using random sampling degrades performance from 25.6\% to 11.9\% (a 13.7\% drop) when Qwen2-VL-7B was used in \textbf{EC}, and from 40.4\% to 11.9\% (a 28.5\% drop) when OS-Atlas-7B was used in \textbf{EC}. These results validate our observation that while MLLMs struggle on high-resolution images, they can still effectively identify coarse yet meaningful region.

\subsection{MLLM Perception}

\subsubsection{Experiment Setting}

\noindent \textbf{Benchmark}\quad
We evaluate our framework on HR-Bench~\cite{hrbench}, a benchmark that consists of multiple-choice questions paired with images at resolution 4K or 8K. HR-Bench consists of two sub-tasks: Fine-grained Single-instance Perception (FSP), which focuses on understanding individual objects, and Fine-grained Cross-instance Perception (FCP), which involves understanding multiple objects and their relationships.

\noindent \textbf{Evaluation Metric}\quad
Following the previous work \cite{hrbench}, we measure accuracy with Cyclic Permutation \cite{cyclic-permutation}, which reorders the answer choices and solves the problem multiple times for each question.

\noindent \textbf{Models}\quad
In the first-stage, \textbf{EC}, we evaluate Qwen2-VL-2B~\cite{qwenvl} and Qwen2-VL-7B~\cite{qwenvl}. For ablation, we add random sampling. In the second stage, \textbf{P}, we conduct experiments using the same models in first-stage, Qwen2-VL-2B and Qwen2-VL-7B.


\begin{table}[t!]
\centering
\resizebox{\columnwidth}{!}{%
\renewcommand{\arraystretch}{1.2}
\begin{tabular}{cccccccc}
\toprule
\multicolumn{2}{c}{\multirow{2}{*}{\textbf{Models}}} & \multicolumn{3}{c}{\textbf{HR-Bench 4K}} & \multicolumn{3}{c}{\textbf{HR-Bench 8K}} \\

& & \textbf{FSP} & \textbf{FCP} & \textbf{Overall} & \textbf{FSP} & \textbf{FCP} & \textbf{Overall} \\

\midrule
\multicolumn{8}{l}{\cellcolor[gray]{0.9}\textbf{\textit{Single-stage Framework}}} \\
\hline
\multicolumn{2}{c}{Qwen2-VL-2B \cite {qwenvl}} & 63.3 & 43.0 & 53.1 & 56.8 & 42.3 & 49.5 \\
\hline
\multicolumn{2}{c}{Qwen2-VL-7B \cite {qwenvl}} & \textbf{71.5} & \textbf{53.5} & \textbf{62.5} & 61.5 & 48.8 & 55.1 \\
\hline
\multicolumn{2}{c}{GPT-4o* \cite{gpt4o}} & 70.0 & 48.0 & 59.0 & \textbf{62.0} & \textbf{49.0} & \textbf{55.5}\\
\midrule
\multicolumn{8}{l}{\cellcolor[gray]{0.9}\textbf{\textit{ECP Framework (Ours)}}}\\
\hline
\multirow{2}{*}{Random Sampling} & Qwen2-VL-2B &  56.8 & 43.5 & 50.1 & 49.0 & 41.3 & 45.1 \\
 & Qwen2-VL-7B &  63.3 & 51.3 & 57.3 & 53.5 & 49.0 & 51.3 \\
\hline
\multirow{2}{*}{Qwen2-VL-2B} & Qwen2-VL-2B & 71.8 & 50.0 & 60.9 & 60.8 & 43.0 & 51.9 \\
& Qwen2-VL-7B & 79.5 & 52.3 & 65.9 & 63.3 & \textbf{49.3} & 56.3 \\
\hline
\multirow{2}{*}{Qwen2-VL-7B} & Qwen2-VL-2B & 77.3 & 49.8 & 63.5 & 69.5 & 45.3 & 57.4 \\
& Qwen2-VL-7B & \textbf{81.0} & \textbf{55.5} & \textbf{68.3} & \textbf{71.8} & 48.8 & \textbf{60.3} \\
\bottomrule
\end{tabular}
}
\caption{Performance comparison of differenct models in single-stage and \textbf{ECP} framework on HR-Bench~\cite{hrbench}. The values marked with * are reported from previous works~\cite{hrbench}.}
\label{tab:main_per}
\vspace{-5mm}
\end{table}

\subsubsection{Experiment Results}
Table~\ref{tab:main_per} presents the results of the conventional single-stage framework and our framework on HR-Bench 4K~\cite{hrbench} and HR-Bench 8K~\cite{hrbench}. 

\noindent \textbf{Single-stage Framework Results}\quad
Among the single-stage framework experiments, which used downsampled images, Qwen2-VL-7B~\cite{qwenvl} achieved the highest overall accuracy of 62.5\% on HR-Bench 4K, while GPT-4o~\cite{gpt4o} achieved the highest overall accuracy of 55.5\% on HR-Bench 8K.

\noindent \textbf{ECP Framework Results}\quad
With our \textbf{ECP} framework, the overall accuracy of Qwen2-VL-7B with the same model for \textbf{EC} increases by 5.8\% on HR-Bench 4K ($62.5\% \rightarrow 68.3\%$) and by 5.2\% on HR-Bench 8K ($55.1\% \rightarrow 60.3\%$). 
In particular, the accuracy on the FSP category, which involves problems related to a single small object, shows a significant improvement achieving +9.5\%, +10.3\% ($71.5\% \rightarrow 81.0\%, 61.5\% \rightarrow 71.8\%$). These improvements in the FSP category indicate that our \textbf{ECP} framework, which selects instruction-relevant areas and uses the corresponding cropped patches, significantly enhances the MLLM's ability to accurately interpret small, single objects.
Additionally, using Qwen2-VL-2B~\cite{qwenvl} instead of Qwen2-VL-7B in \textbf{EC} or \textbf{P} results in a slight decrease in overall accuracy on HR-Bench 4K and HR-Bench 8K. Nevertheless, using \textbf{EC} still outperforms single-stage framework, indicating that the Qwen2-VL-2B model is capable of selecting important regions within the image. This suggests that, depending on available computational resources, using Qwen2-VL-7B in \textbf{EC} can be used for higher performance, while the Qwen2-VL-2B offers a more resource-efficient alternative.

\begin{figure}[]
    \centering
    \includegraphics[width=\linewidth]{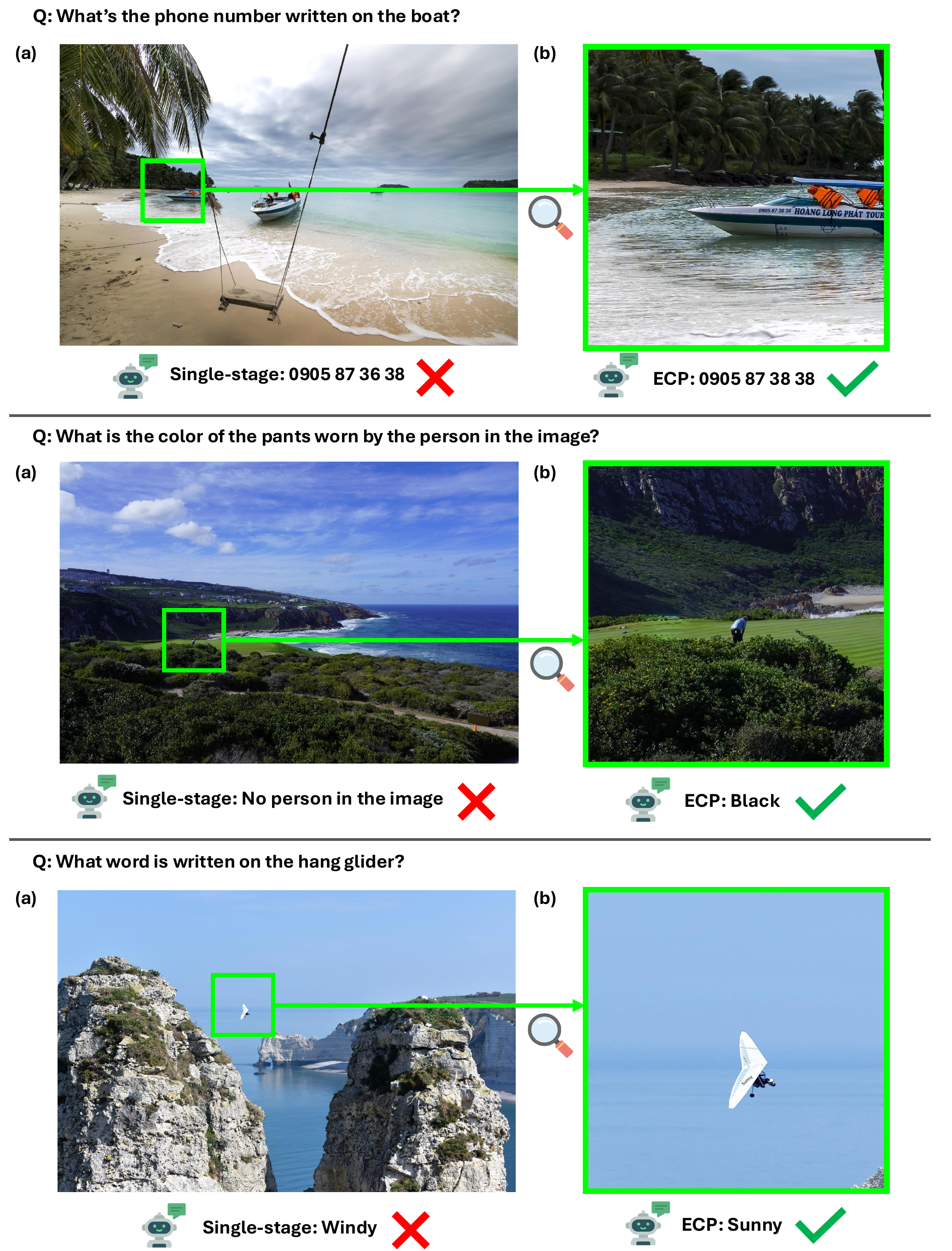}
    \caption{Qualitative results on HR-Bench 8K~\cite{hrbench} comparing \textbf{(a)} conventional single-stage framework and \textbf{(b)} the proposed \textbf{ECP} framework.}
    \label{fig:fig3}
    \vspace{-5mm}
\end{figure}

\begin{figure}[]
    \centering
    \includegraphics[width=\linewidth]{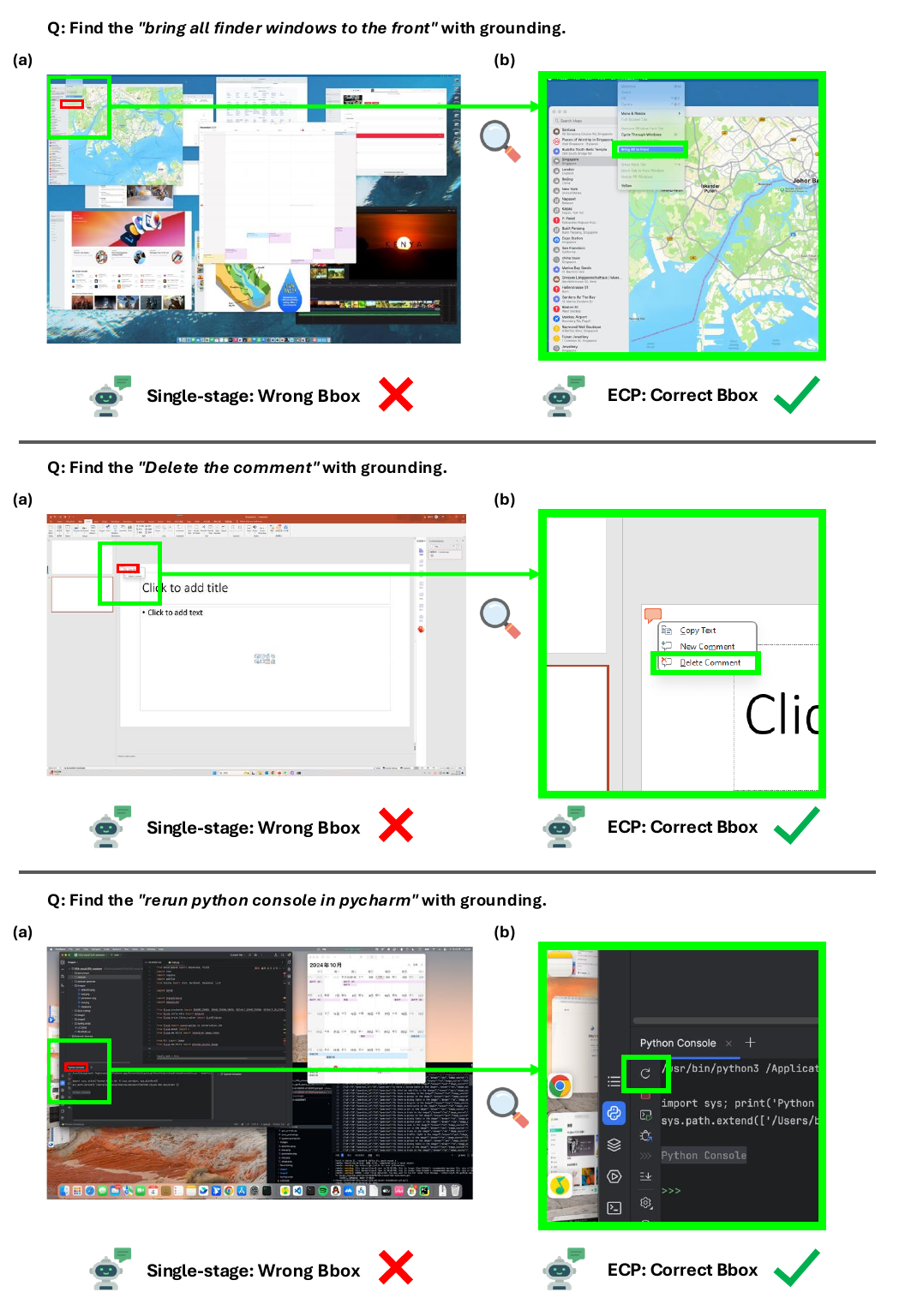}  
    \caption{Qualitative results on ScreenSpot-Pro~\cite{screenspot-pro} comparing \textbf{(a)} conventional single-stage framework and \textbf{(b)} the proposed \textbf{ECP} framework.}
    \label{fig:fig4}
    \vspace{-5mm}
\end{figure}

\noindent \textbf{Qualitative Results}\quad
Figure~\ref{fig:fig3} and Figure~\ref{fig:fig4} show qualitative results of the conventional single-stage framework and our \textbf{ECP} framework on HR-Bench 8K and Screenspot-Pro. On HR-Bench 8K, the single-stage framework fails to answer questions about small objects. In contrast, our \textbf{ECP} framework enables the MLLM to first locate the relevant object and then answer correctly using the cropped image. Similarly, on Screenspot-Pro, while the single-stage framework fails to ground the target correctly, our \textbf{ECP} framework identifies its approximate location and provides accurate answers using the cropped image.


\noindent\textbf{Ablation}\quad
For ablation, when applying random sampling in \textbf{EC}, the performance dropped significantly, resulting in an overall accuracy that was even lower than that of the single-stage framework. For Qwen2-VL-2B, performance decreases from 63.5\% to 50.1\% (a 13.4\% drop) on HR-Bench 4K and from 57.4\% to 45.1\% (a 12.3\% drop) on HR-Bench 8K when using Qwen2-VL-7B in \textbf{EC}. The same degradation also observed for Qwen2-VL-7B. These results demonstrate that while MLLMs struggle with understanding small objects in high-resolution images, they are still capable of roughly identifying meaningful regions.

\section{Conclusion}
\label{sec:Conclusion}

We propose \textbf{E}xtract \textbf{C}andidate then \textbf{P}redict (\textbf{ECP}), a novel two-stage framework where the MLLM first extracts a candidate from a downsampled high-resolution image and then makes prediction based on the candidate. Unlike prior methods, our framework is training-free and task-agnostic, making it widely applicable to various high-resolution vision-language tasks. We validate the effectiveness of \textbf{ECP} on 4K GUI grounding and 4K, 8K MLLM perception. While our framework is inherently training-free and task-agnostic, the current evaluation focuses on a limited set of downstream tasks and benchmarks. Expanding the experimental coverage to a broader range of tasks and comparing against additional baselines would provide a more comprehensive validation of \textbf{ECP}’s generality. Furthermore, we leave the exploration of different image cropping strategies and ablation studies across various MLLMs as promising avenues to better understand the generalizability and robustness of \textbf{ECP}.

{
    \small
    \bibliographystyle{ieeenat_fullname}
    \bibliography{main}

\begin{thebibliography}{24}
\providecommand{\natexlab}[1]{#1}
\providecommand{\url}[1]{\texttt{#1}}
\expandafter\ifx\csname urlstyle\endcsname\relax
  \providecommand{\doi}[1]{doi: #1}\else
  \providecommand{\doi}{doi: \begingroup \urlstyle{rm}\Url}\fi

\bibitem[Bai et~al.(2023)Bai, Bai, Yang, Wang, Tan, Wang, Lin, Zhou, and Zhou]{qwenvl}
Jinze Bai, Shuai Bai, Shusheng Yang, Shijie Wang, Sinan Tan, Peng Wang, Junyang Lin, Chang Zhou, and Jingren Zhou.
\newblock Qwen-vl: A versatile vision-language model for understanding, localization, text reading, and beyond.
\newblock \emph{arXiv preprint arXiv:2308.12966}, 2023.

\bibitem[Brown et~al.(2020)Brown, Mann, Ryder, Subbiah, Kaplan, Dhariwal, Neelakantan, Shyam, Sastry, Askell, et~al.]{gpt3}
Tom Brown, Benjamin Mann, Nick Ryder, Melanie Subbiah, Jared~D Kaplan, Prafulla Dhariwal, Arvind Neelakantan, Pranav Shyam, Girish Sastry, Amanda Askell, et~al.
\newblock Language models are few-shot learners.
\newblock \emph{Advances in neural information processing systems}, 33:\penalty0 1877--1901, 2020.

\bibitem[Caron et~al.(2021)Caron, Touvron, Misra, J{\'e}gou, Mairal, Bojanowski, and Joulin]{dino}
Mathilde Caron, Hugo Touvron, Ishan Misra, Herv{\'e} J{\'e}gou, Julien Mairal, Piotr Bojanowski, and Armand Joulin.
\newblock Emerging properties in self-supervised vision transformers.
\newblock In \emph{Proceedings of the IEEE/CVF international conference on computer vision}, pages 9650--9660, 2021.

\bibitem[Chen et~al.(2024)Chen, Wu, Wang, Su, Chen, Xing, Zhong, Zhang, Zhu, Lu, et~al.]{internvl}
Zhe Chen, Jiannan Wu, Wenhai Wang, Weijie Su, Guo Chen, Sen Xing, Muyan Zhong, Qinglong Zhang, Xizhou Zhu, Lewei Lu, et~al.
\newblock Internvl: Scaling up vision foundation models and aligning for generic visual-linguistic tasks.
\newblock In \emph{Proceedings of the IEEE/CVF conference on computer vision and pattern recognition}, pages 24185--24198, 2024.

\bibitem[Cheng et~al.(2024)Cheng, Sun, Chu, Xu, Li, Zhang, and Wu]{seeclick}
Kanzhi Cheng, Qiushi Sun, Yougang Chu, Fangzhi Xu, Yantao Li, Jianbing Zhang, and Zhiyong Wu.
\newblock Seeclick: Harnessing gui grounding for advanced visual gui agents.
\newblock \emph{arXiv preprint arXiv:2401.10935}, 2024.

\bibitem[Chowdhery et~al.(2023)Chowdhery, Narang, Devlin, Bosma, Mishra, Roberts, Barham, Chung, Sutton, Gehrmann, et~al.]{palm}
Aakanksha Chowdhery, Sharan Narang, Jacob Devlin, Maarten Bosma, Gaurav Mishra, Adam Roberts, Paul Barham, Hyung~Won Chung, Charles Sutton, Sebastian Gehrmann, et~al.
\newblock Palm: Scaling language modeling with pathways.
\newblock \emph{Journal of Machine Learning Research}, 24\penalty0 (240):\penalty0 1--113, 2023.

\bibitem[Guo et~al.(2024)Guo, Xu, Yao, Cui, Ni, Ge, Chua, Liu, and Huang]{llava-uhd}
Zonghao Guo, Ruyi Xu, Yuan Yao, Junbo Cui, Zanlin Ni, Chunjiang Ge, Tat-Seng Chua, Zhiyuan Liu, and Gao Huang.
\newblock Llava-uhd: an lmm perceiving any aspect ratio and high-resolution images, 2024.

\bibitem[Hurst et~al.(2024)Hurst, Lerer, Goucher, Perelman, Ramesh, Clark, Ostrow, Welihinda, Hayes, Radford, et~al.]{gpt4o}
Aaron Hurst, Adam Lerer, Adam~P Goucher, Adam Perelman, Aditya Ramesh, Aidan Clark, AJ Ostrow, Akila Welihinda, Alan Hayes, Alec Radford, et~al.
\newblock Gpt-4o system card.
\newblock \emph{arXiv preprint arXiv:2410.21276}, 2024.

\bibitem[Li et~al.(2025)Li, Meng, Lin, Luo, Tian, Ma, Huang, and Chua]{screenspot-pro}
Kaixin Li, Ziyang Meng, Hongzhan Lin, Ziyang Luo, Yuchen Tian, Jing Ma, Zhiyong Huang, and Tat-Seng Chua.
\newblock Screenspot-pro: Gui grounding for professional high-resolution computer use, 2025.

\bibitem[Liu et~al.(2023)Liu, Li, Wu, and Lee]{llava}
Haotian Liu, Chunyuan Li, Qingyang Wu, and Yong~Jae Lee.
\newblock Visual instruction tuning.
\newblock \emph{Advances in neural information processing systems}, 36:\penalty0 34892--34916, 2023.

\bibitem[Liu et~al.(2024)Liu, You, Han, Wang, Zhai, Liu, Tao, Huang, He, and Yang]{infimm-hd}
Haogeng Liu, Quanzeng You, Xiaotian Han, Yiqi Wang, Bohan Zhai, Yongfei Liu, Yunzhe Tao, Huaibo Huang, Ran He, and Hongxia Yang.
\newblock Infimm-hd: A leap forward in high-resolution multimodal understanding.
\newblock \emph{arXiv preprint arXiv:2403.01487}, 2024.

\bibitem[Luo et~al.(2024)Luo, Zhou, Zhang, Zheng, Sun, and Ji]{llava-hr}
Gen Luo, Yiyi Zhou, Yuxin Zhang, Xiawu Zheng, Xiaoshuai Sun, and Rongrong Ji.
\newblock Feast your eyes: Mixture-of-resolution adaptation for multimodal large language models.
\newblock \emph{arXiv preprint arXiv:2403.03003}, 2024.

\bibitem[Ma et~al.(2024)Ma, Wang, Sun, Lin, Zhou, Ji, and Ji]{inf-llava}
Yiwei Ma, Zhibin Wang, Xiaoshuai Sun, Weihuang Lin, Qiang Zhou, Jiayi Ji, and Rongrong Ji.
\newblock Inf-llava: Dual-perspective perception for high-resolution multimodal large language model.
\newblock \emph{arXiv preprint arXiv:2407.16198}, 2024.

\bibitem[Radford et~al.(2021)Radford, Kim, Hallacy, Ramesh, Goh, Agarwal, Sastry, Askell, Mishkin, Clark, et~al.]{clip}
Alec Radford, Jong~Wook Kim, Chris Hallacy, Aditya Ramesh, Gabriel Goh, Sandhini Agarwal, Girish Sastry, Amanda Askell, Pamela Mishkin, Jack Clark, et~al.
\newblock Learning transferable visual models from natural language supervision.
\newblock In \emph{International conference on machine learning}, pages 8748--8763. PmLR, 2021.

\bibitem[Rolih et~al.(2024)Rolih, Ameln, Vaidya, and Akcay]{dc}
Bla{\v{z}} Rolih, Dick Ameln, Ashwin Vaidya, and Samet Akcay.
\newblock Divide and conquer: High-resolution industrial anomaly detection via memory efficient tiled ensemble.
\newblock In \emph{Proceedings of the IEEE/CVF Conference on Computer Vision and Pattern Recognition}, pages 3866--3875, 2024.

\bibitem[Team et~al.(2024)Team, Georgiev, Lei, Burnell, Bai, Gulati, Tanzer, Vincent, Pan, Wang, et~al.]{gemini15}
Gemini Team, Petko Georgiev, Ving~Ian Lei, Ryan Burnell, Libin Bai, Anmol Gulati, Garrett Tanzer, Damien Vincent, Zhufeng Pan, Shibo Wang, et~al.
\newblock Gemini 1.5: Unlocking multimodal understanding across millions of tokens of context.
\newblock \emph{arXiv preprint arXiv:2403.05530}, 2024.

\bibitem[Touvron et~al.(2019)Touvron, Vedaldi, Douze, and J{\'e}gou]{fixres}
Hugo Touvron, Andrea Vedaldi, Matthijs Douze, and Herv{\'e} J{\'e}gou.
\newblock Fixing the train-test resolution discrepancy.
\newblock \emph{Advances in neural information processing systems}, 32, 2019.

\bibitem[Touvron et~al.(2023)Touvron, Lavril, Izacard, Martinet, Lachaux, Lacroix, Rozi{\`e}re, Goyal, Hambro, Azhar, et~al.]{llama}
Hugo Touvron, Thibaut Lavril, Gautier Izacard, Xavier Martinet, Marie-Anne Lachaux, Timoth{\'e}e Lacroix, Baptiste Rozi{\`e}re, Naman Goyal, Eric Hambro, Faisal Azhar, et~al.
\newblock Llama: Open and efficient foundation language models.
\newblock \emph{arXiv preprint arXiv:2302.13971}, 2023.

\bibitem[Wang et~al.(2024{\natexlab{a}})Wang, Bai, Tan, Wang, Fan, Bai, Chen, Liu, Wang, Ge, et~al.]{qwen}
Peng Wang, Shuai Bai, Sinan Tan, Shijie Wang, Zhihao Fan, Jinze Bai, Keqin Chen, Xuejing Liu, Jialin Wang, Wenbin Ge, et~al.
\newblock Qwen2-vl: Enhancing vision-language model's perception of the world at any resolution.
\newblock \emph{arXiv preprint arXiv:2409.12191}, 2024{\natexlab{a}}.

\bibitem[Wang et~al.(2024{\natexlab{b}})Wang, Ding, Zeng, Zhou, Shen, Luo, and Tao]{hrbench}
Wenbin Wang, Liang Ding, Minyan Zeng, Xiabin Zhou, Li Shen, Yong Luo, and Dacheng Tao.
\newblock Divide, conquer and combine: A training-free framework for high-resolution image perception in multimodal large language models.
\newblock \emph{arXiv preprint arXiv:2408.15556}, 2024{\natexlab{b}}.

\bibitem[Wu et~al.(2024)Wu, Wu, Xu, Wang, Sun, Jia, Cheng, Ding, Chen, Liang, et~al.]{os-atlas}
Zhiyong Wu, Zhenyu Wu, Fangzhi Xu, Yian Wang, Qiushi Sun, Chengyou Jia, Kanzhi Cheng, Zichen Ding, Liheng Chen, Paul~Pu Liang, et~al.
\newblock Os-atlas: A foundation action model for generalist gui agents.
\newblock \emph{arXiv preprint arXiv:2410.23218}, 2024.

\bibitem[Zhang et~al.(2025)Zhang, Li, Fei, Yuan, Wu, Ji, Loy, and Yan]{omgllava}
Tao Zhang, Xiangtai Li, Hao Fei, Haobo Yuan, Shengqiong Wu, Shunping Ji, Chen~Change Loy, and Shuicheng Yan.
\newblock Omg-llava: Bridging image-level, object-level, pixel-level reasoning and understanding.
\newblock \emph{Advances in Neural Information Processing Systems}, 37:\penalty0 71737--71767, 2025.

\bibitem[Zheng et~al.(2023)Zheng, Zhou, Meng, Zhou, and Huang]{cyclic-permutation}
Chujie Zheng, Hao Zhou, Fandong Meng, Jie Zhou, and Minlie Huang.
\newblock Large language models are not robust multiple choice selectors.
\newblock \emph{arXiv preprint arXiv:2309.03882}, 2023.

\bibitem[Zhu et~al.(2024)Zhu, Zhu, Chen, Xu, Tang, and Wang]{mrovseg}
Yuanbing Zhu, Bingke Zhu, Zhen Chen, Huan Xu, Ming Tang, and Jinqiao Wang.
\newblock Mrovseg: Breaking the resolution curse of vision-language models in open-vocabulary semantic segmentation.
\newblock \emph{arXiv preprint arXiv:2408.14776}, 2024.

\end{thebibliography}
}


\end{document}